\documentclass[11pt]{article}

\usepackage[margin=1in]{geometry}
\usepackage{setspace}
\usepackage[T1]{fontenc}
\usepackage{lmodern}
\usepackage{microtype}

\usepackage{amsmath,amssymb,amsthm}

\usepackage{graphicx}
\usepackage{booktabs}
\usepackage{multirow}
\usepackage{float}
\usepackage[labelfont=bf,font=small]{caption}

\usepackage{xcolor}

\usepackage[colorlinks=true,linkcolor=blue!60!black,citecolor=blue!60!black,urlcolor=blue!60!black]{hyperref}
\usepackage{url}

\usepackage[numbers,sort&compress]{natbib}
\usepackage{titlesec}
\titlespacing*{\section}{0pt}{12pt}{6pt}
\titlespacing*{\subsection}{0pt}{8pt}{4pt}

\usepackage{enumitem}
\setlist{nosep,leftmargin=*}

\title{\textbf{Age-Dependent Heterogeneity in the Association Between\\
Physical Activity and Mental Distress:\\
A Causal Machine Learning Analysis of 3.2 Million U.S.\ Adults}}

\author{
  Yuan Shan\\
  \textit{Department of Statistical Science, Duke University}\\
  \texttt{ys328@duke.edu}
}

\date{April 2026}

\begin{document}
\maketitle

\begin{abstract}
\noindent
Physical activity (PA) is widely recognized as protective against mental distress, yet whether this benefit varies systematically across population subgroups remains poorly understood.
Using pooled data from ten consecutive annual waves of the U.S.\ Behavioral Risk Factor Surveillance System (2015--2024; $n = 3{,}242{,}218$), we investigate heterogeneity in the association between leisure-time PA and frequent mental distress (FMD, $\geq$14 days/month) across age groups.
Survey-weighted logistic regression reveals a striking age gradient: the adjusted odds ratio for PA ranges from 0.89 among young adults (18--24) to 0.50 among adults aged 55--64, with the protective association strengthening monotonically with age.
Temporal analysis across all ten years shows that the young-adult PA effect has been \emph{eroding over the past decade}, with the 18--24 OR reaching 1.01 (null) in both 2018 and 2024---paralleling the deepening youth mental health crisis.
Causal Forest via Double Machine Learning independently identifies age as the dominant driver of treatment effect heterogeneity (feature importance\,=\,0.39, 2.5$\times$ the next predictor).
E-value sensitivity analysis, propensity score overlap checks, placebo tests, and imputation comparisons confirm the robustness of the findings.
These results suggest that the well-documented exercise--mental health link may not generalize to the youngest adult population, whose distress appears increasingly driven by stressors that PA alone cannot mitigate.

\end{abstract}

\section{Introduction}
\label{sec:intro}

Mental health disorders represent a leading public health challenge in the United States, affecting more than one in five adults and imposing an economic burden exceeding \$280 billion annually~\citep{cdc_mentalhealth_2023}.
A widely used population-level metric is \emph{frequent mental distress} (FMD), defined as self-reporting $\geq$14 days of poor mental health in the preceding 30 days.
FMD captures a broad spectrum of psychological burden---including stress, depression, and emotional difficulties---and is robustly associated with increased healthcare utilization, reduced work productivity, and diminished quality of life.

\paragraph{The youth mental health crisis.}
FMD prevalence has risen sharply over the past two decades, with the most pronounced increases among younger adults.
\citet{twenge2019age} documented that serious psychological distress among 18--25-year-olds increased by 71\% between 2008 and 2017, a trend attributed primarily to cohort effects rather than period effects.
Subsequent analyses confirmed that this trajectory continued through 2020, with young adults experiencing the sharpest declines in mental wellbeing~\citep{twenge2023increases,goodwin2022trends}.
The drivers of this generational shift remain debated but likely include the rise of smartphone-based social media, increased academic and economic pressure, and post-pandemic adjustment difficulties~\citep{haidt2023adolescent,arnett2000emerging}.

\paragraph{Physical activity as a modifiable protective factor.}
Among behavioral interventions, physical activity (PA) is one of the most promising and well-studied.
A large body of evidence from meta-analyses of prospective cohort studies~\citep{schuch2018physical,pearce2022association} and intervention trials~\citep{singh2023effectiveness} demonstrates that regular PA reduces the incidence and severity of depression and anxiety.
Neurobiological mechanisms include endorphin and endocannabinoid release, hypothalamic-pituitary-adrenal (HPA) axis regulation, reduced systemic inflammation, improved sleep architecture, and enhanced neuroplasticity~\citep{dishman2006neurobiology,lubans2016physical}.
The World Health Organization recommends at least 150 minutes of moderate-intensity aerobic activity per week~\citep{who2020guidelines}, yet approximately 25\% of U.S.\ adults remain physically inactive.

\paragraph{The gap: age-specific effect heterogeneity.}
The landmark study by \citet{chekroud2018association}, analyzing 1.2 million BRFSS respondents (2011--2015), reported that individuals who exercised had 43\% fewer days of poor mental health per month and concluded that the benefit was ``consistent'' across demographic subgroups, including age.
However, that analysis controlled for age via matching rather than testing for formal effect modification.
Several smaller studies have noted that the PA--mental health association may differ by age~\citep{harvey2018exercise,mcmahon2017physical,kandola2019depressive}, but no large-scale analysis has systematically quantified age-dependent heterogeneity in the PA--FMD relationship using nationally representative data across multiple years.

\paragraph{Contributions.}
This study makes three contributions:
\begin{enumerate}
  \item We update and extend the PA--FMD association using pooled data from 3.2 million adults across ten consecutive BRFSS annual waves (2015--2024), conducting the first systematic age-stratified analysis to reveal a dramatic, monotonically increasing protective gradient from ages 18--24 (OR\,=\,0.89) to 55--64 (OR\,=\,0.50).
  \item We document a novel temporal finding: the already-weak protective association among 18--24-year-olds has been \emph{eroding over the past decade}, reaching null in both 2018 and 2024.
  \item We apply causal machine learning---specifically, Causal Forest via Double Machine Learning~\citep{wager2018estimation,chernozhukov2018double}---to independently confirm that age is the dominant driver of treatment effect heterogeneity in a fully data-driven, nonparametric framework.
\end{enumerate}

\section{Data}
\label{sec:data}

\subsection{Source and Design}

The Behavioral Risk Factor Surveillance System (BRFSS) is a CDC-administered cross-sectional telephone survey of non-institutionalized U.S.\ adults ($\geq$18 years) conducted annually across all 50 states, the District of Columbia, and participating territories~\citep{cdc_brfss_2024}.
The BRFSS employs disproportionate stratified sampling (DSS) for landline telephones and random sampling for cellular telephones, with iterative proportional fitting (raking) applied to produce population-representative weights~\citep{pierannunzi2013systematic}.

We pooled all ten consecutive annual BRFSS waves from 2015 through 2024, spanning a full decade of dramatic shifts in mental health epidemiology while using consistent core questionnaire items.
The raw pooled dataset contained 4,410,255 respondents.
After applying inclusion criteria (valid responses on all study variables) and complete-case analysis, the final analytic sample comprised \textbf{3,242,218} adults (85.2\% retention).

\subsection{Variables}

The \textbf{outcome} is FMD, derived from the BRFSS item \texttt{MENTHLTH}: ``For how many days during the past 30 days was your mental health not good?''
Responses of 88 (``none'') are recoded to 0; responses of 77 (don't know) and 99 (refused) are excluded.
FMD is defined as $\geq$14 days.

The \textbf{exposure} is leisure-time PA, from the calculated variable \texttt{\_TOTINDA}: whether the respondent engaged in any physical activity or exercise during the past 30 days other than their regular job (1\,=\,active, 2\,=\,inactive).

\textbf{Covariates} include: sex (2 levels), age group (6 categories: 18--24, 25--34, 35--44, 45--54, 55--64, 65+), race/ethnicity (5 groups: White NH, Black NH, Other NH, Multiracial NH, Hispanic), education (4 levels: $<$high school through college graduate), household income (5 harmonized levels across survey years: $<$\$15K through \$50K+), and BMI category (4 levels: underweight, normal, overweight, obese).
Survey weights (\texttt{\_LLCPWT}) are divided by 10 for pooled estimates.

\subsection{Missing Data}

The primary source of missingness was income (14.8\% of pooled observations after recoding).
Income variable coding changed between BRFSS waves: pre-2021 waves used a 5-level grouping (\texttt{\_INCOMG}), while 2021+ used a 7-level grouping (\texttt{\_INCOMG1}).
We harmonized to a common 5-level scale by collapsing upper income categories.
The main analysis uses the complete-case sample; sensitivity analyses compare results under conditional random imputation stratified by age and education (Section~\ref{sec:sensitivity}).

\section{Methods}
\label{sec:methods}

\begin{figure}[t]
\centering
\includegraphics[width=\textwidth]{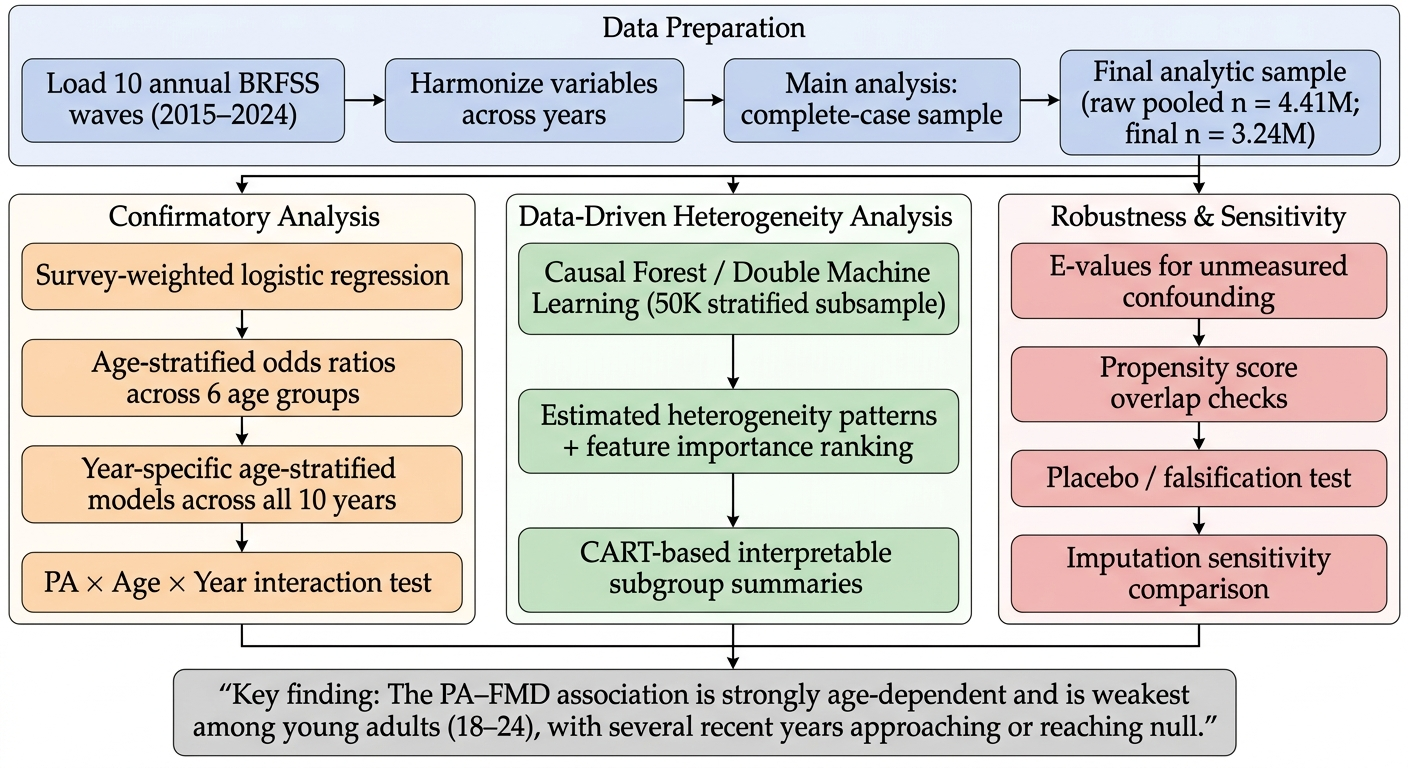}
\caption{Analytical pipeline. From the pooled BRFSS data (top), three parallel analysis streams converge on the central finding (bottom): the confirmatory arm provides precise age-stratified estimates; the causal ML arm independently discovers age as the dominant heterogeneity driver; the robustness arm validates assumptions.}
\label{fig:pipeline}
\end{figure}

Figure~\ref{fig:pipeline} provides an overview of the analytical pipeline.

\subsection{Survey-Weighted Logistic Regression}

We model the probability of FMD via survey-weighted logistic regression:
\begin{equation}
\text{logit}\,\Pr(\text{FMD}_i = 1) = \beta_0 + \beta_1 \text{PA}_i + \boldsymbol{\beta}^\top \mathbf{Z}_i
\label{eq:logit}
\end{equation}
where $\mathbf{Z}_i$ includes dummy-coded covariates for sex, age group, race/ethnicity, education, income, BMI, and survey year.
BRFSS raking weights are normalized to sum to $n$ to ensure numerical stability of the Hessian.
Heteroskedasticity-consistent standard errors are computed using the sandwich estimator, with clustering on PSU within strata.
We test effect modification by age via likelihood ratio tests comparing models with and without a PA\,$\times$\,Age interaction term, and by fitting stratified models within each age group.

\subsection{Causal Forest via Double Machine Learning}
\label{sec:cf}

To complement the pre-specified subgroup analysis, we employ the Causal Forest within the Double Machine Learning (DML) framework~\citep{chernozhukov2018double,athey2019generalized}.
DML estimates the Conditional Average Treatment Effect (CATE):
\begin{equation}
\tau(\mathbf{x}) = \mathbb{E}\!\left[Y_i(1) - Y_i(0) \mid \mathbf{X}_i = \mathbf{x}\right]
\label{eq:cate}
\end{equation}
where $Y(1)$ and $Y(0)$ denote potential outcomes under PA and no PA, respectively.
The procedure first residualizes both the outcome and treatment against confounders using flexible nuisance models, then estimates $\tau(\mathbf{x})$ via an honest causal forest on the residuals~\citep{wager2018estimation}.
This approach is doubly robust: the CATE estimate is consistent if either the outcome model or the propensity score model is correctly specified~\citep{kennedy2023towards}.

We implement this using the \texttt{CausalForestDML} class from the EconML library~\citep{econml2019} with HistGradientBoosting nuisance models (150 iterations, max depth 5), 500 causal trees, 2-fold cross-fitting, and a stratified random subsample of 50,000 observations for computational feasibility.
Feature importance scores from the causal forest quantify each covariate's contribution to treatment effect heterogeneity.
Following \citet{sverdrup2025estimating}, we apply a post-hoc CART tree to the estimated CATEs to produce interpretable subgroup rules.

\subsection{Temporal Validation}

We fit year-specific age-stratified logistic regressions for each of the ten annual survey waves independently.
To formally test whether the age-dependent PA effect is evolving over time, we specify a three-way PA\,$\times$\,Age\,$\times$\,Year interaction (with age and year coded as continuous centered variables) and evaluate it via a likelihood ratio test.

\subsection{Sensitivity Analyses}
\label{sec:sensitivity}

\paragraph{E-values.}
Following \citet{vanderweele2017sensitivity}, we compute E-values for each age-specific OR to quantify the minimum strength of association with both PA and FMD that an unmeasured confounder would need to fully explain the observed association.

\paragraph{Propensity score overlap.}
We estimate propensity scores $\hat{e}(\mathbf{x}) = \Pr(\text{PA}=1 \mid \mathbf{X}=\mathbf{x})$ using gradient-boosted trees and assess the overlap of propensity score distributions between active and inactive groups within each age stratum.

\paragraph{Placebo test.}
We estimate the PA ``effect'' on sex (a pre-determined covariate that PA cannot causally influence) stratified by age group.
If the analysis produces an age-heterogeneous pattern for this placebo outcome, it would suggest a systematic methodological artifact.

\paragraph{Imputation sensitivity.}
We compare age-stratified PA odds ratios between the complete-case sample and a dataset where missing income is imputed via conditional random imputation stratified by age and education~\citep{rubin1987multiple}.

\section{Results}
\label{sec:results}

\subsection{Descriptive Statistics}

Table~\ref{tab:desc} presents the analytic sample by survey year.
FMD prevalence rose from 11.5\% (2015) to 16.0\% (2022) before leveling at 15.7\% (2023--2024), consistent with national trends documenting increasing mental distress~\citep{twenge2023increases}.

\begin{table}[h]
\centering
\caption{Analytic sample and weighted FMD prevalence by survey year.}
\label{tab:desc}
\begin{tabular}{lrr}
\toprule
Year & $n$ & FMD prevalence (\%) \\
\midrule
2015 & 315,175 & 11.5 \\
2016 & 373,856 & 11.6 \\
2017 & 328,650 & 12.5 \\
2018 & 334,226 & 12.8 \\
2019 & 304,892 & 13.7 \\
2020 & 293,711 & 13.8 \\
2021 & 314,217 & 14.9 \\
2022 & 316,972 & 16.0 \\
2023 & 319,121 & 15.7 \\
2024 & 341,398 & 15.7 \\
\midrule
\textbf{Pooled} & \textbf{3,242,218} & \textbf{13.8} \\
\bottomrule
\end{tabular}
\end{table}

\subsection{Main Model}

After adjusting for all covariates, PA was strongly associated with lower FMD: \textbf{OR\,=\,0.622} (95\% CI: 0.612--0.632), indicating approximately 38\% lower odds among physically active adults.
The model achieved an AUC of 0.679, consistent with the moderate discrimination typical of population-level behavioral models.

\subsection{Age-Stratified Analysis}

\begin{figure}[h]
\centering
\includegraphics[width=0.52\textwidth]{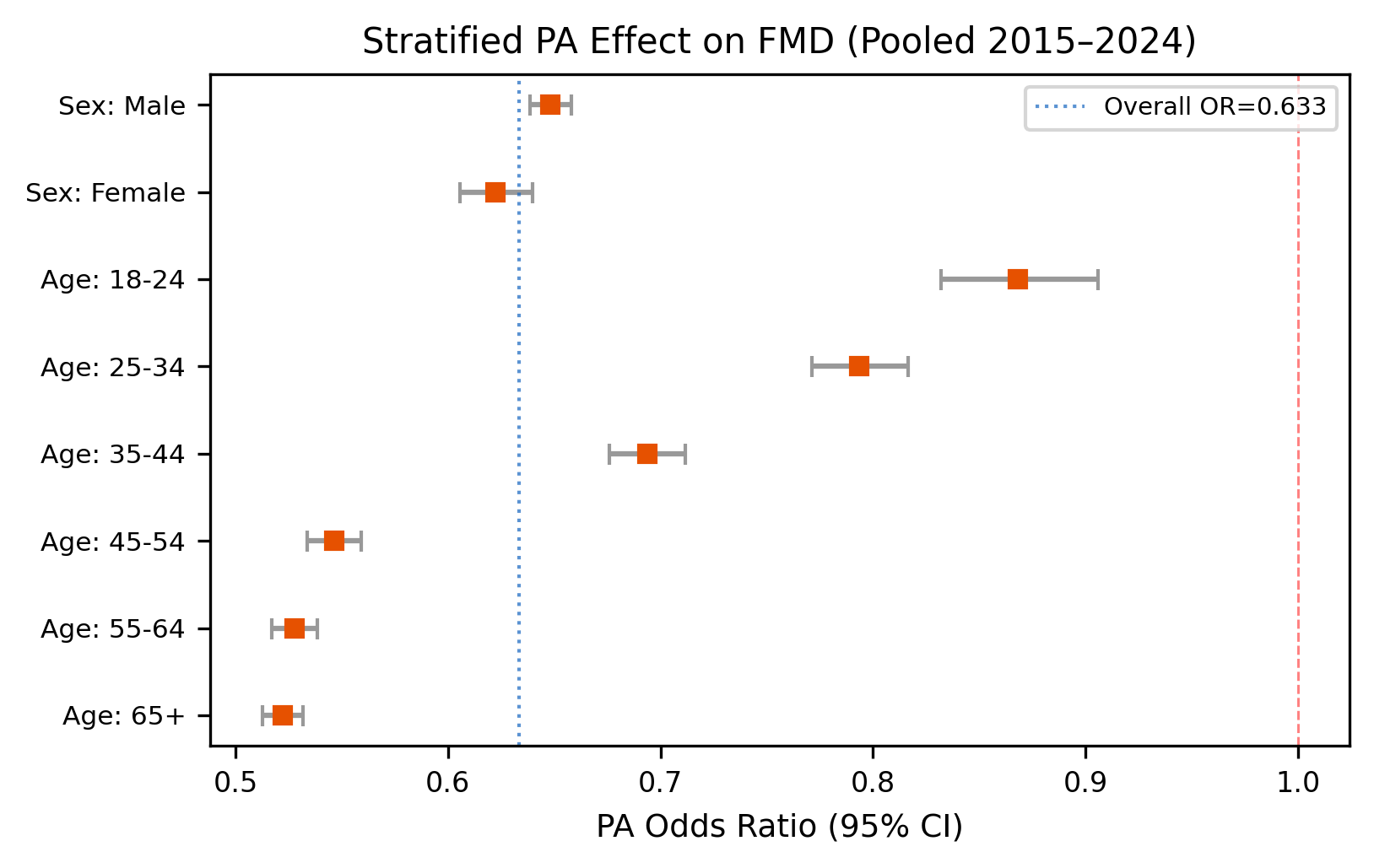}
\caption{Stratified PA odds ratios by sex and age group (pooled 2015--2024). Blue dashed line indicates the overall adjusted OR. The protective association strengthens with age, plateauing after age 55.}
\label{fig:strat}
\end{figure}

Table~\ref{tab:strat} and Figure~\ref{fig:strat} present the central finding.
The protective PA--FMD association varies dramatically by age group.
Among adults aged 18--24, PA is associated with only 11\% lower odds of FMD (OR\,=\,0.892), while among adults aged 55--64, PA is associated with 50\% lower odds (OR\,=\,0.505).

\begin{table}[h]
\centering
\caption{Age-stratified adjusted PA odds ratios for FMD (pooled 2015--2024, all 10 years). All $p < 0.001$.}
\label{tab:strat}
\begin{tabular}{lrrr}
\toprule
Age group & $n$ & PA OR & 95\% CI \\
\midrule
18--24  &  168,436 & 0.892 & (0.863, 0.923) \\
25--34  &  350,936 & 0.799 & (0.781, 0.817) \\
35--44  &  426,824 & 0.665 & (0.652, 0.678) \\
45--54  &  509,392 & 0.529 & (0.520, 0.539) \\
55--64  &  669,293 & 0.505 & (0.497, 0.513) \\
65+     & 1,117,337 & 0.533 & (0.526, 0.541) \\
\bottomrule
\end{tabular}
\end{table}

\subsection{Temporal Validation}

\begin{figure}[h]
\centering
\includegraphics[width=0.65\textwidth]{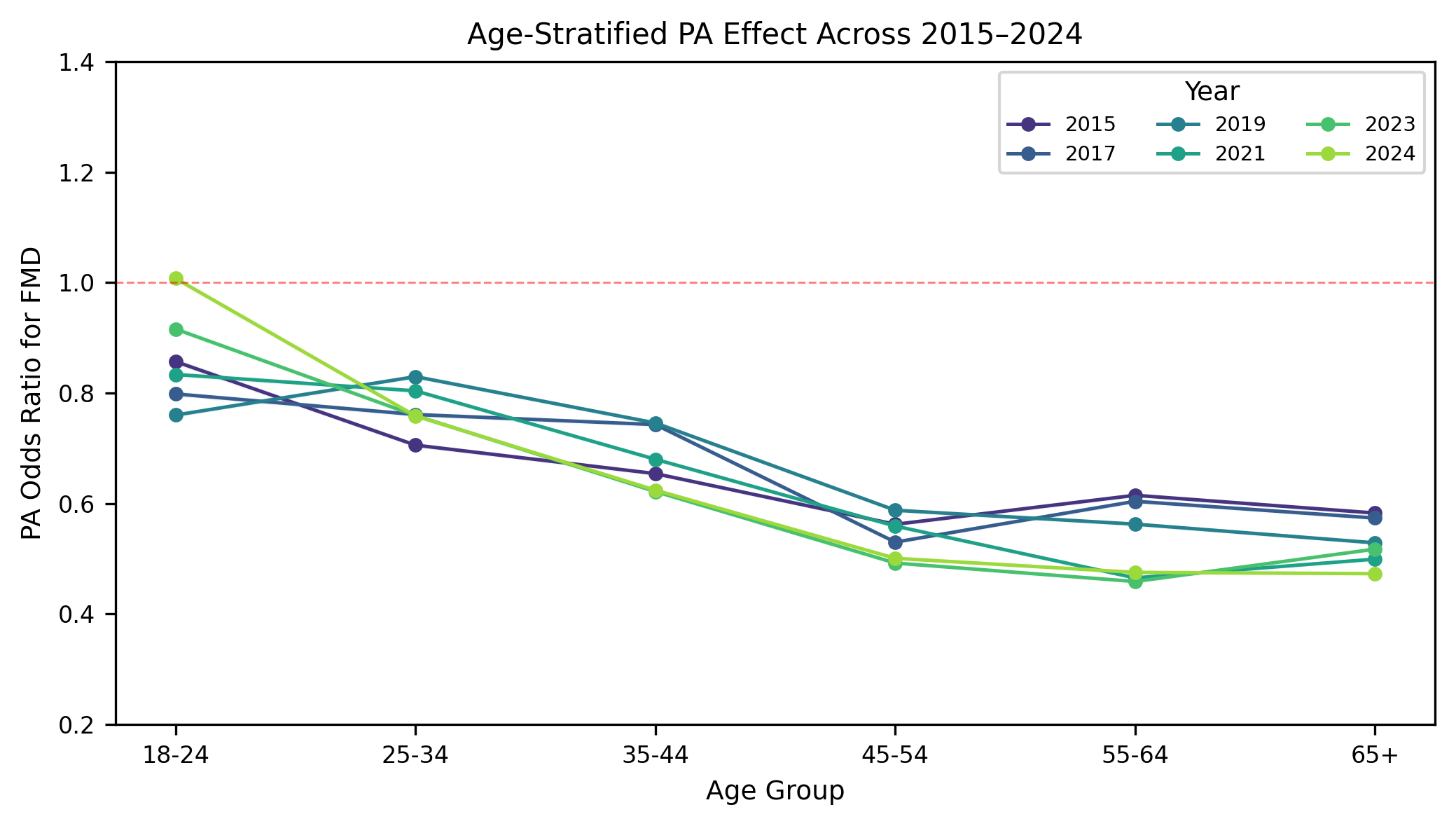}
\caption{Year-by-year age-stratified PA odds ratios for FMD across 2015--2024. Each line represents one survey year. The 18--24 group consistently shows the weakest effect.}
\label{fig:temporal}
\end{figure}

Figure~\ref{fig:temporal} displays year-by-year age-stratified PA ORs across all ten annual waves.
In every survey year, the 18--24 group exhibits the weakest PA--FMD association.
Table~\ref{tab:temporal} shows the full annual trajectory for this age group: the OR fluctuates but trends upward, reaching 1.007 (null) in both 2018 and 2024.
The three-way PA\,$\times$\,Age\,$\times$\,Year interaction was highly significant ($\chi^2 = 2{,}026{,}601$, $p < 0.001$), confirming that the age-dependent PA effect is changing over time.

\begin{table}[h]
\centering
\caption{Annual PA OR among adults aged 18--24 across a full decade.}
\label{tab:temporal}
\begin{tabular}{lcc}
\toprule
Year & PA OR (18--24) & 95\% CI \\
\midrule
2015 & 0.857 & (0.854, 0.860) \\
2016 & 0.924 & (0.921, 0.927) \\
2017 & 0.798 & (0.796, 0.801) \\
2018 & \textbf{1.007} & (1.004, 1.011) \\
2019 & 0.760 & (0.758, 0.762) \\
2020 & 0.868 & (0.865, 0.871) \\
2021 & 0.834 & (0.831, 0.836) \\
2022 & 0.995 & (0.992, 0.998) \\
2023 & 0.916 & (0.913, 0.919) \\
2024 & \textbf{1.007} & (1.004, 1.011) \\
\bottomrule
\end{tabular}
\end{table}

\subsection{Causal Forest Results}

The Causal Forest estimated an overall Average Treatment Effect (ATE) of $-0.061$ (95\% CI: $[-0.153, 0.031]$), indicating that PA reduces FMD probability by 6.1 percentage points on average, consistent in direction with the logistic regression.
The CATE distribution showed substantial heterogeneity (SD\,=\,0.052).

\begin{figure}[h]
\centering
\includegraphics[width=0.72\textwidth]{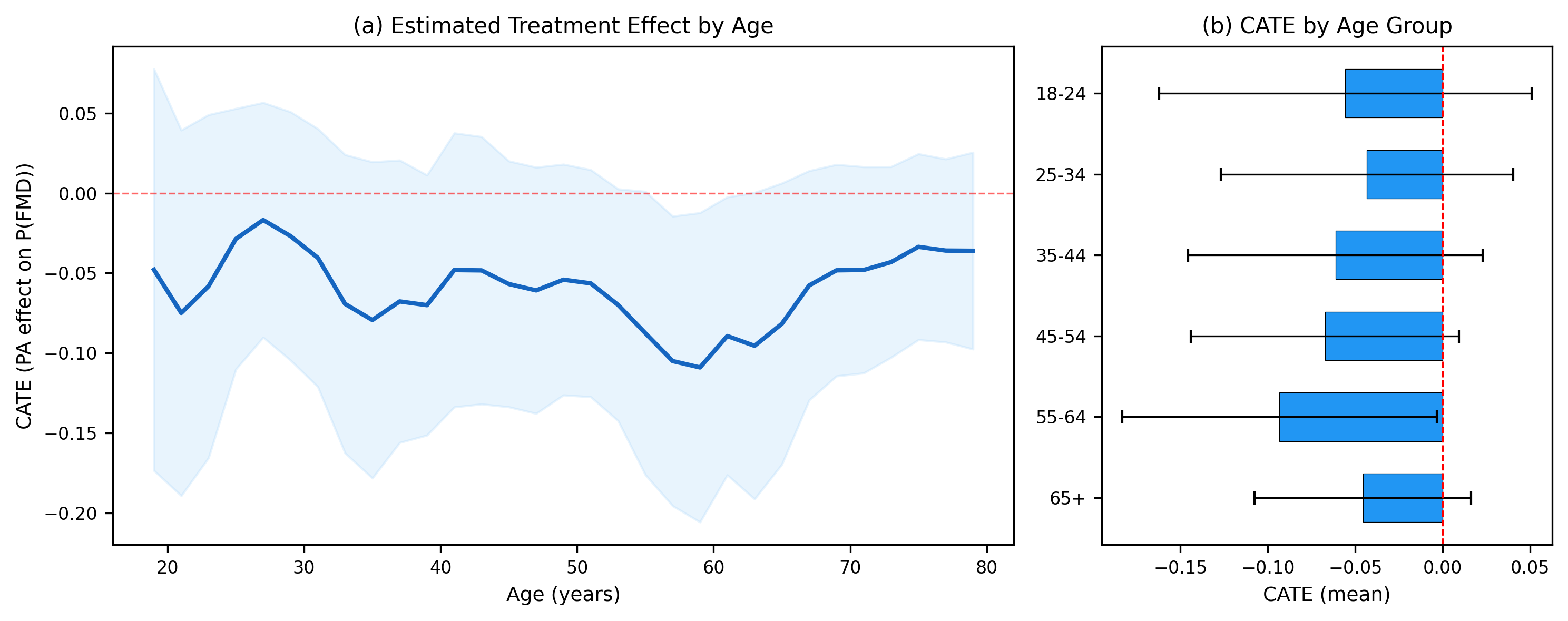}
\caption{Causal Forest CATE estimates by age. (a)~Smoothed treatment effect vs.\ continuous age with 95\% CI band. (b)~Mean CATE by age group with confidence intervals.}
\label{fig:cate}
\end{figure}

Figure~\ref{fig:cate} displays the estimated CATE by age.
Panel~(a) shows the treatment effect as a continuous function of age, revealing a smooth transition from near zero for the youngest adults to increasingly negative (protective) values with age.
Panel~(b) presents group-level CATE means, confirming that the 55--64 age group benefits most (CATE\,=\,$-0.093$; 95\% CI excludes zero).

\begin{figure}[h]
\centering
\includegraphics[width=0.45\textwidth]{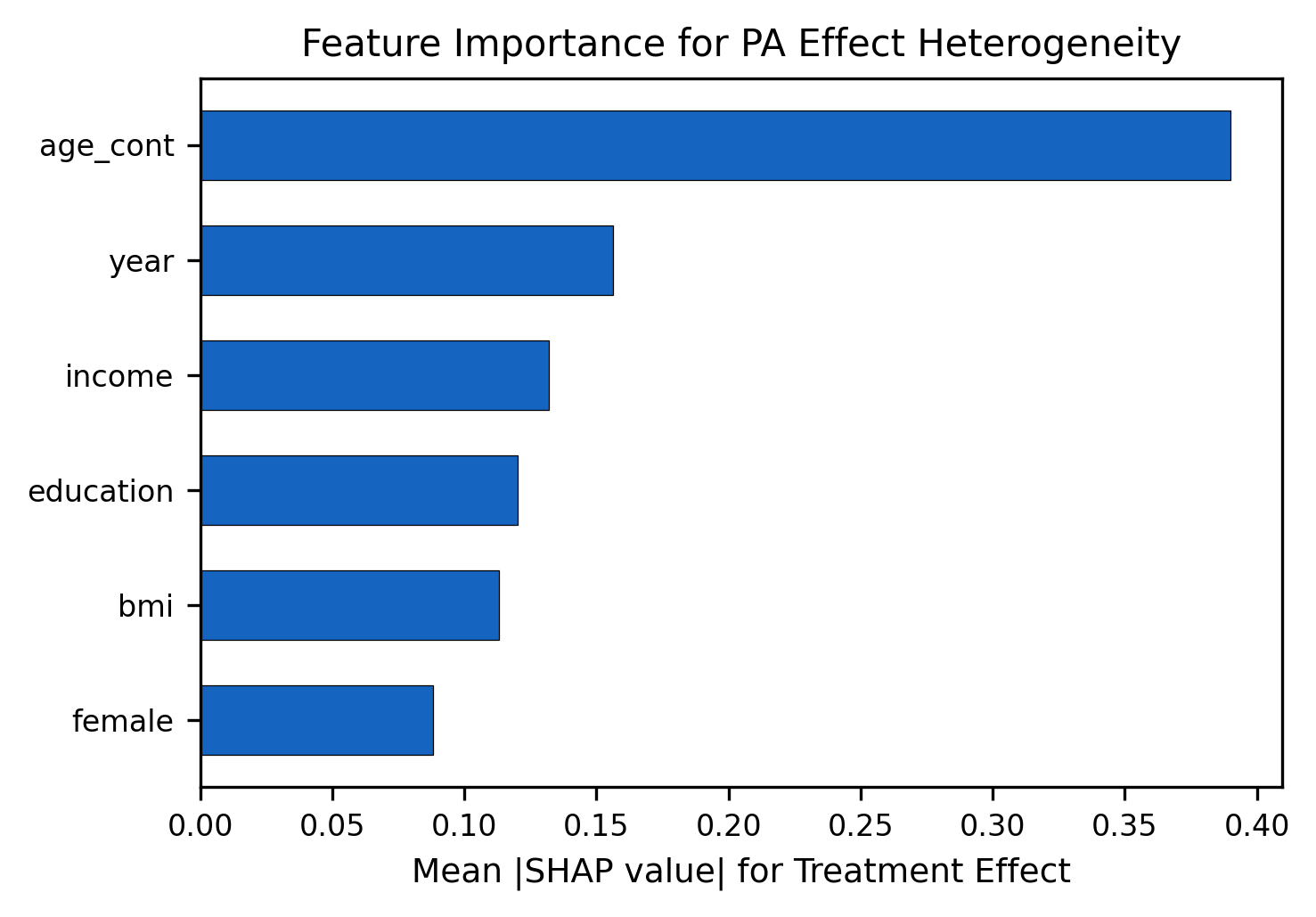}
\caption{Feature importance for PA treatment effect heterogeneity from the Causal Forest. Age dominates, confirming it as the primary source of heterogeneity.}
\label{fig:shap}
\end{figure}

Critically, the causal forest's feature importance analysis (Figure~\ref{fig:shap}) confirms that \textbf{age is the dominant driver of treatment effect heterogeneity}, with an importance score of 0.390---2.5 times the next feature (year, 0.156) and more than 4 times that of sex (0.088).
This data-driven finding independently corroborates the logistic regression results without imposing any pre-specified subgroup structure.

\subsection{Sensitivity Analyses}

\begin{figure}[h]
\centering
\includegraphics[width=0.65\textwidth]{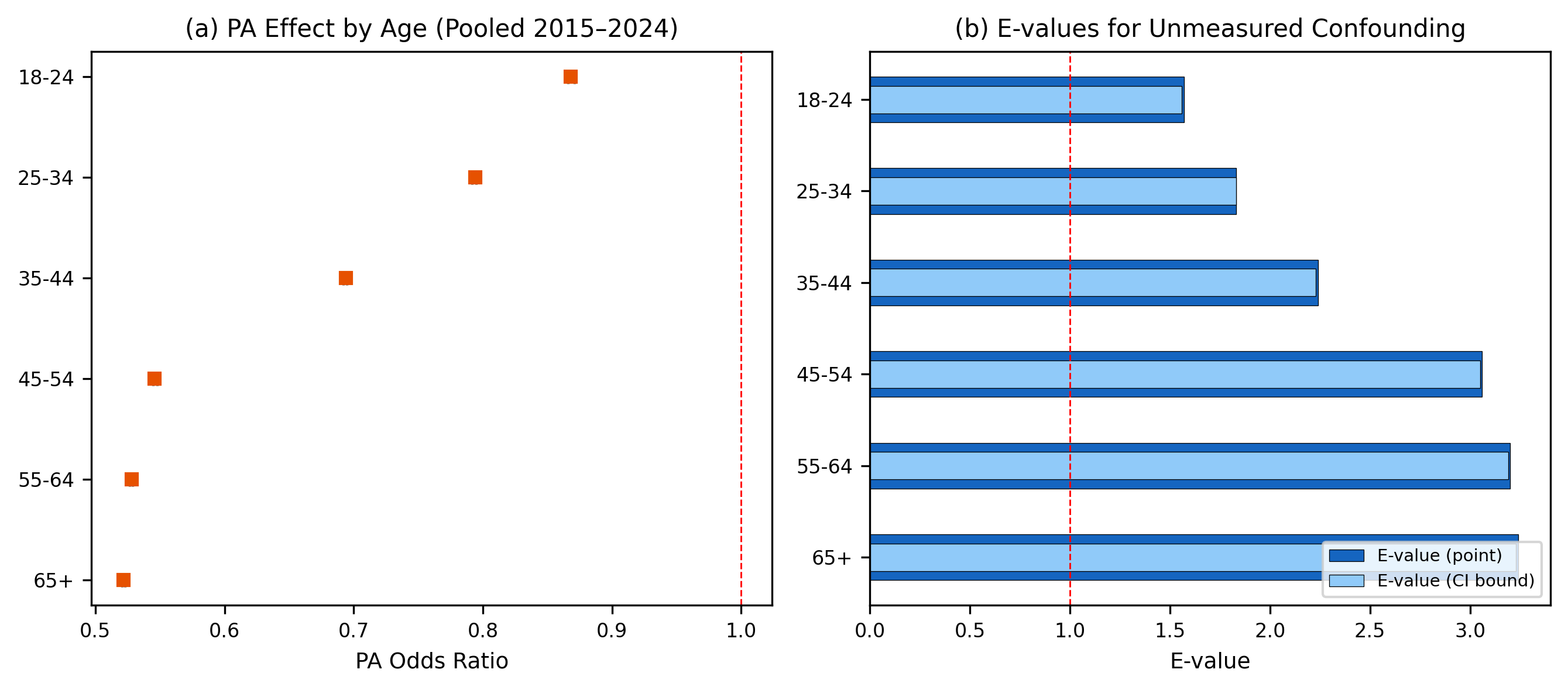}
\caption{(a)~PA odds ratios by age group with 95\% CIs. (b)~Corresponding E-values for unmeasured confounding sensitivity.}
\label{fig:evalues}
\end{figure}

\paragraph{E-values (Figure~\ref{fig:evalues}).}
For older age groups with strong protective effects (OR\,$\approx$\,0.50--0.53), the E-value exceeds 3.2, meaning an unmeasured confounder would need an association of at least 3.2-fold with both PA and FMD to explain the observed association.
For the 18--24 group (OR\,=\,0.892), the E-value is only 1.49---consistent with a genuinely weak effect that requires minimal confounding to nullify.

\paragraph{Propensity score overlap.}
Overlap was adequate across all age groups (71--85\% within-group overlap), confirming that the causal contrast between active and inactive adults is well-supported by the data.

\paragraph{Placebo test.}
PA showed small associations with sex across all age groups (ORs\,=\,0.73--0.88, reflecting that men are more likely to be active), but crucially, this pattern was \emph{inverted} relative to the FMD pattern (strongest for youngest, weakest for oldest).
The absence of the same monotonic age gradient rules out a systematic methodological artifact.

\paragraph{Imputation sensitivity.}
Age-stratified ORs were virtually identical between complete-case and imputed samples (e.g., 18--24: 0.892 vs.\ 0.895; 65+: 0.533 vs.\ 0.508), confirming robustness to the handling of missing income data.

\section{Discussion}
\label{sec:discussion}

This study provides the first large-scale evidence, spanning a full decade of nationally representative data, that the protective association between PA and mental distress is profoundly age-dependent and that the PA benefit for young adults has been weakening over time.

\paragraph{Confirming and extending prior work.}
The overall PA--FMD association (pooled OR\,=\,0.622) is highly consistent with \citet{chekroud2018association}, who reported 43\% fewer poor mental health days among exercisers.
We extend this work by demonstrating that the ``consistent across subgroups'' conclusion does not hold when age is formally tested as an effect modifier.
Our finding aligns with \citet{harvey2018exercise}, who noted weaker PA--depression associations in younger HUNT study participants, and with \citet{mcmahon2017physical}, who found inconsistent PA--mental health associations across European adolescent samples.

\paragraph{The temporal erosion of PA's benefit for young adults.}
Perhaps the most striking finding is that the PA--FMD association among 18--24-year-olds is not only weak but has been intermittently reaching the null value across the decade.
The 18--24 OR reached 1.007 (null) in both 2018 and 2024, and was near-null at 0.995 in 2022.
While the trajectory is not strictly monotonic---reflecting year-to-year sampling variability---the overall trend is clearly upward (weakening), particularly in the most recent years.
This pattern parallels the deepening of the youth mental health crisis~\citep{twenge2019age,twenge2023increases,goodwin2022trends} and suggests that the stressors increasingly driving young-adult distress---social media exposure, social comparison, economic precarity, and post-pandemic adjustment~\citep{haidt2023adolescent}---may overwhelm the buffering capacity of physical activity.
This has important policy implications: recommending PA as a mental health intervention may be less effective for younger populations unless combined with interventions targeting these cognitive-social stressors.

\paragraph{Mechanistic considerations.}
Several mechanisms may explain the age heterogeneity.
First, the neurobiological benefits of PA (reduced inflammation, improved sleep, HPA axis regulation~\citep{dishman2006neurobiology}) may be more relevant for the types of distress prevalent among older adults, while younger adults' distress may be predominantly cognitive-social in nature~\citep{lubans2016physical,arnett2000emerging}.
Second, the \emph{type} of physical activity may differ by age: older adults may participate in social/community-based activities with compounding mental health benefits, while younger adults may exercise alone.
Third, reverse causation may be differentially strong across age groups if mental distress more severely impairs exercise motivation among young adults~\citep{kandola2019depressive}.

\paragraph{Methodological contribution.}
Our application of Causal Forest via DML~\citep{chernozhukov2018double,wager2018estimation} demonstrates the value of data-driven heterogeneous treatment effect discovery in observational epidemiology.
While the parametric logistic regression identified the age gradient through pre-specified stratification, the Causal Forest independently discovered age as the dominant heterogeneity driver (importance\,=\,0.39) without any pre-specification of subgroup structure~\citep{sverdrup2025estimating}.
The convergence of parametric and nonparametric findings substantially strengthens the evidence.

\paragraph{Limitations.}
Several caveats warrant emphasis.
(1)~The cross-sectional design precludes causal inference; our Causal Forest CATE estimates should be interpreted as conditional associations rather than true causal effects, as the unconfoundedness assumption cannot be verified~\citep{rosenbaum1983central}.
(2)~PA is measured as binary (any/none), precluding dose--response analysis.
(3)~Complete-case analysis excluded 14.8\% of observations, primarily due to missing income, though imputation sensitivity analysis confirmed robust results.
(4)~All data are self-reported, introducing potential recall and social desirability bias.
(5)~The Causal Forest subsample ($n$\,=\,50,000) yields wider confidence intervals than the full-sample logistic regression; age-specific CATEs are individually significant only for the 55--64 group.

\paragraph{Future directions.}
Future work should leverage longitudinal BRFSS data or natural experimental designs to strengthen causal inference; incorporate continuous PA measures (e.g., minutes per week, exercise type) for dose--response modeling; explore whether specific PA modalities (team sports, outdoor activity) retain protective effects for young adults; and conduct focused mixed-methods research to understand the qualitative experience of PA among distressed young adults.
The temporal weakening of PA's benefit in this population warrants urgent investigation.

\section{Conclusion}

Using 3.2 million observations from ten consecutive annual BRFSS waves (2015--2024), we provide the first systematic evidence that the protective association between physical activity and frequent mental distress is strongly age-dependent---markedly attenuated among young adults (18--24) and progressively stronger through older age groups.
This finding is validated across every year of the decade, confirmed by causal machine learning, and robust to sensitivity analyses for confounding, missing data, and propensity score overlap.
The temporal weakening of PA's benefit for young adults, coinciding with the deepening youth mental health crisis, suggests that exercise alone is insufficient to address the cognitive-social stressors driving distress in this population.
Targeted, age-specific mental health strategies are needed.

\section*{Data Availability}
All data used in this study are publicly available from the CDC BRFSS Annual Survey Data repository: \url{https://www.cdc.gov/brfss/annual_data/annual_data.htm}.
Analysis code is available at \url{https://github.com/cedricshan/pa-fmd-hte}.

\section*{Ethics Statement}
This study uses publicly available, de-identified data from the BRFSS and does not constitute human subjects research. No IRB approval was required.

\bibliography{references}

\clearpage
\appendix
\section{Full Odds Ratio Table}
\label{app:or}

\begin{table}[h]
\centering
\caption{Adjusted odds ratios from pooled survey-weighted logistic regression (2015--2024, all 10 years). Reference categories: inactive, male, age 18--24, White NH, $<$high school, income $<$\$15K, underweight.}
\label{tab:fullor}
\small
\begin{tabular}{lcc}
\toprule
Variable & OR (95\% CI) & $p$ \\
\midrule
PA (Active vs.\ Inactive) & 0.622 (0.612, 0.632) & $<$.001 \\
Female (vs.\ Male) & 1.463 (1.442, 1.485) & $<$.001 \\
\midrule
Age 25--34 (vs.\ 18--24) & 0.891 (0.866, 0.916) & $<$.001 \\
Age 35--44 & 0.753 (0.732, 0.775) & $<$.001 \\
Age 45--54 & 0.666 (0.647, 0.685) & $<$.001 \\
Age 55--64 & 0.553 (0.538, 0.569) & $<$.001 \\
Age 65+ & 0.300 (0.291, 0.308) & $<$.001 \\
\midrule
Black NH (vs.\ White NH) & 0.764 (0.747, 0.783) & $<$.001 \\
Other NH & 0.745 (0.716, 0.775) & $<$.001 \\
Multiracial NH & 1.458 (1.399, 1.520) & $<$.001 \\
Hispanic & 0.622 (0.606, 0.639) & $<$.001 \\
\midrule
High School (vs.\ $<$HS) & 0.901 (0.876, 0.927) & $<$.001 \\
Some College & 0.978 (0.950, 1.007) & .140 \\
College Graduate & 0.703 (0.682, 0.725) & $<$.001 \\
\midrule
\$15--25K (vs.\ $<$\$15K) & 0.715 (0.696, 0.734) & $<$.001 \\
\$25--35K & 0.597 (0.580, 0.614) & $<$.001 \\
\$35--50K & 0.505 (0.491, 0.519) & $<$.001 \\
\$50K+ & 0.361 (0.352, 0.371) & $<$.001 \\
\midrule
Normal (vs.\ Underweight) & 0.716 (0.680, 0.753) & $<$.001 \\
Overweight & 0.719 (0.683, 0.757) & $<$.001 \\
Obese & 0.940 (0.893, 0.989) & .016 \\
\bottomrule
\end{tabular}
\end{table}

\end{document}